\documentclass[journal]{IEEEtran}
\IEEEoverridecommandlockouts
\usepackage{cite}
\usepackage{amsmath,amssymb,amsfonts}
\usepackage{graphicx}
\usepackage{textcomp}
\usepackage{color}
\usepackage{xcolor}
\usepackage[linesnumbered,ruled]{algorithm2e}
\usepackage{booktabs}
\usepackage{array}
\usepackage{algpseudocode}
\usepackage{amsmath}
\usepackage{tabularx}
\usepackage{makecell}
\newcolumntype{P}[1]{>{\centering\arraybackslash}p{#1}}
\newcolumntype{M}[1]{>{\centering\arraybackslash}m{#1}}
\usepackage{amssymb}
\usepackage{CJK}
\usepackage{multirow}
\usepackage{listings}
\usepackage{xcolor}
\usepackage{bm,graphicx,multirow,multicol,bm,bbm,psfrag,subfigure,color,mdframed,wasysym,subeqnarray}
\usepackage{hyphenat}
\usepackage{enumitem} 
\usepackage{soul}
\usepackage[colorlinks,
linkcolor=blue,
anchorcolor=black,
citecolor=blue,
urlcolor=black
]{hyperref}
\soulregister\cite7
\soulregister\ref7 
\soulregister\eqref7 
\soulregister\pageref7 

\begin{document}
	
	\title{
		A Novel Planning Framework for Complex Flipping Manipulation of Multiple Mobile Manipulators
		
	}
	
	\author{Wenhang~Liu, Meng~Ren, Kun Song, Michael~Yu~Wang,~\IEEEmembership{Fellow,~IEEE}, and Zhenhua~Xiong,~\IEEEmembership{Member,~IEEE}
		
		\thanks{
			This work was supported in part by the National Natural Science Foundation of China (U1813224) and MoE Key Lab of Artificial Intelligence, AI Institute, Shanghai Jiao Tong University, China. \textit{(Corresponding author: Zhenhua Xiong.)}
		}
		\thanks{
			Wenhang~Liu, Meng~Ren, Kun Song, and Zhenhua~Xiong are with the School of Mechanical Engineering, State Key Laboratory of Mechanical System and Vibration, Shanghai Jiao Tong University, Shanghai, China (e-mail: liuwenhang@sjtu.edu.cn; meng\_ren@sjtu.edu.cn; coldtea@sjtu.edu.cn; mexiong@sjtu.edu.cn).
		}
		\thanks{
			Michael~Yu~Wang is with the Department of Mechanical and Aerospace Engineering, Hong Kong University of Science and Technology, Clear Water Bay, Hong Kong, and also with the School of Engineering, Great Bay University, Guangdong, China (e-mail: mywang@ust.hk; mywang@gbu.edu.cn).
		}
	}
	
	\mark{Thhis work has been submitted to the IEEE for possible publication. Copyright may be transferred without notice, after which this version may no longer be accessible.}
	
	\maketitle

	\begin{abstract}
		
		During complex object manipulation, manipulator systems often face the configuration disconnectivity problem due to closed-chain constraints.
		Although regrasping can be adopted to get a piecewise connected manipulation, it is a challenging problem to determine whether there is a planning result without regrasping.
		To address this problem, a novel planning framework is proposed for multiple mobile manipulator systems.
		Coordinated platform motions and regrasping motions are proposed to enhance configuration connectivity. 
		Given the object trajectory and the grasping pose set, the planning framework includes three steps.
		First, inverse kinematics for each mobile manipulator is verified along the given trajectory based on different grasping poses.
		Coverable trajectory segments are determined for each robot for a specific grasping pose.
		Second, the trajectory choice problem is formulated into a set cover problem, by which we can quickly determine whether the manipulation can be completed without regrasping or with the minimal regrasping number.
		Finally, the motions of each mobile manipulator are planned with the assigned trajectory segments using existing methods.
		Both simulations and experimental results show the performance of the planner in complex flipping manipulation.
		Additionally, the proposed planner can greatly extend the adaptability of multiple mobile manipulator systems in complex manipulation tasks.

	\end{abstract}
	

	\section{Introduction}
	\IEEEPARstart{C}{ompared} to a single and powerful robot, Multiple Robot Systems (MRS) are well-suited to handle intricate tasks since they 
	exhibit enhanced robustness, stability, and efficiency\cite{dorigo2020reflections}. 
	Hence, the versatility of MRS has gained increasing attention and become a focal point of interest in various applications, such as assembly\cite{suarez2018can}\cite{dogar2015multi}, exploration\cite{arm2023scientific}, and transporting\cite{kennel2024payload}\cite{kim2022learning}.
	Among different forms of MRS, Multiple Mobile Manipulator Systems (MMMS) stand out for their integration of mobility and manipulation capabilities, making them particularly attractive for industrial automation\cite{fiore2024general}.
	In automated production, MMMS can accomplish complex processes by collaborative manipulation, for example, they are applied to manipulate industrial components in \cite{dogar2015multi}.
	When it comes to large objects or complex manipulations, MMMS are even irreplaceable.
	
	\begin{figure}[t]
		\includegraphics[width=\columnwidth]{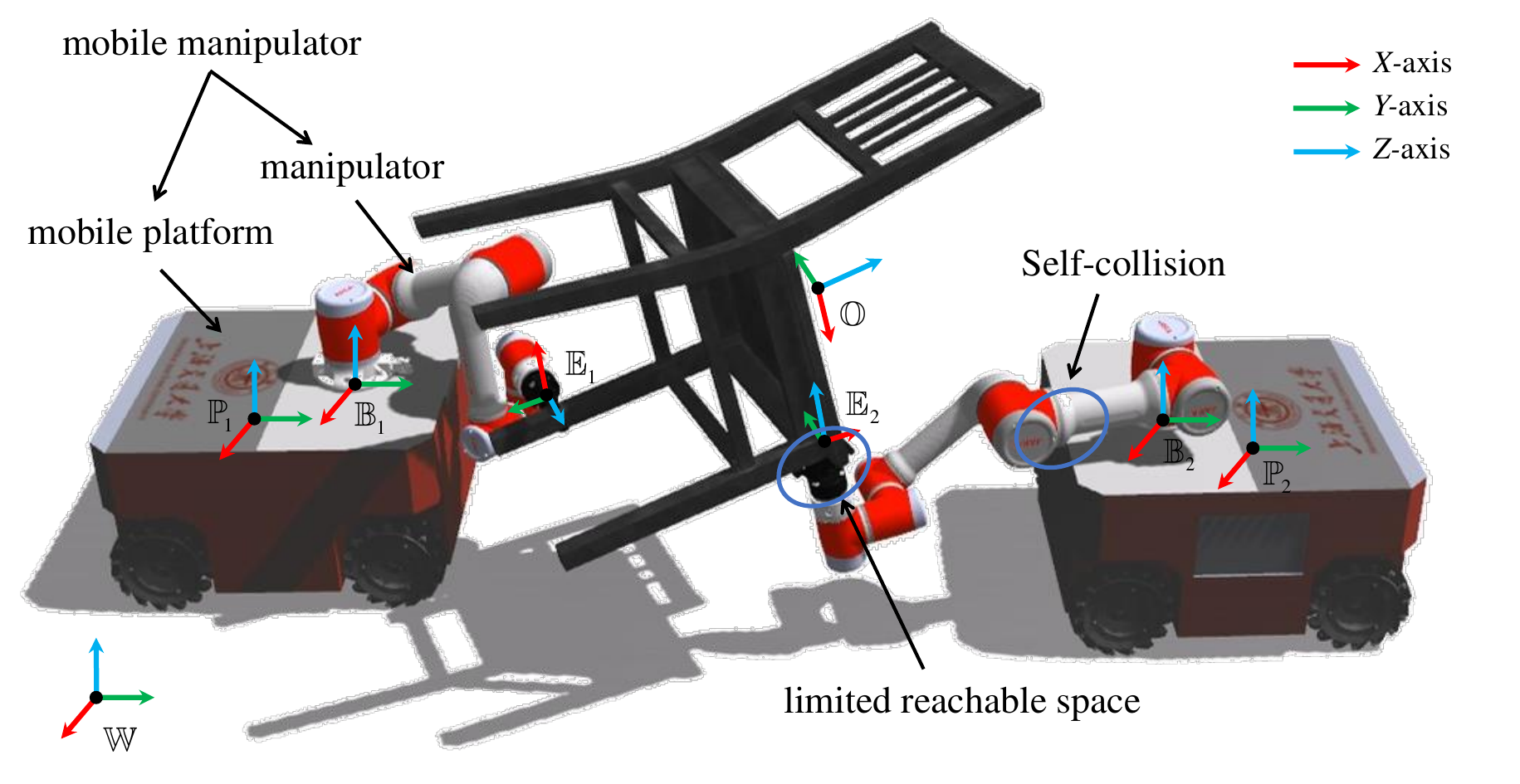}
		\centering
		\caption{
			The right side manipulator suffers the configuration disconnectivity due to unreachable space and self-collision, resulting in the inability to continue manipulating the object along the desired trajectory.
		}
		\label{MMMS}       
	\end{figure}
	
	Although MMMS possess great potential in object manipulation, there are still challenges for efficient coordination of MRS.
	In terms of manipulation planning, the biggest challenge arises from the configuration disconnectivity of manipulators caused by closed-chain constraints\cite{xian2017closed}, which require that the grasping pose of each manipulator remains fixed during the manipulation.
	A major issue arises when, for a given object trajectory, the configuration of a manipulator $\bm{q}_m$ is near the boundary of its valid configuration space $\mathcal{C}_{\textit{free}}$.
	An example is shown in Fig.~\ref{MMMS}.
	At this point, the current grasping pose of the right manipulator needs to change in order to avoid self-collision and continue the object manipulation process.

	\subsection{Related Work}
	There are generally two types of methods to deal with closed-chain constraints.
	One type involves considering the configuration of each robot individually.
	In \cite{cortes2002random}\cite{bordalba2020randomized}, probabilistic roadmaps were utilized to efficiently determine feasible configurations for robots.
	In \cite{zhang2021task}, the global planner first explored the low-dimensional decomposed task space regions of each robot, and then the local planner computed paths in the high-dimensional constrained configuration space.
	Another type is to consider the overall configuration of the system.
	In \cite{jiao2015transportation}, three modes were predefined according to the objects and closed-chain constraints, based on which, a decision-making method was proposed to facilitate the selection of object transportation and movement direction.
	In \cite{alonso2017multi}, the closed-chain constraints were transformed into the problem of finding the optimal formation for the whole system.
	Planning the overall configuration can conveniently facilitate the integration with other tasks, such as obstacle avoidance.
	
	The aforementioned works enable MMMS to plan feasible configurations under closed-chain constraints and successfully manipulate objects, as demonstrated through experiments or simulations\cite{cortes2002random}\cite{zhang2021task,jiao2015transportation,alonso2017multi}.
	However, it is worth noting that these works primarily address tasks related to object transportation.
	The large translational movements mostly rely on the mobility of mobile platforms.
	The manipulator with a fixed grasping pose only performs small coordinated motions if necessary.
	Due to the small range of motion, the manipulator usually does not suffer the problem of configuration disconnectivity caused by potential collision or joint limits under the closed-chain constraints \cite{xian2017closed}.
	
	However, in more complex manipulation tasks such as flipping, each manipulator requires extensive motions, and the issue of configuration disconnectivity is prominent.
	In \cite{xian2017closed}, configuration disconnectivity of stationary-mounted manipulators was addressed in flipping manipulations.
	A method termed IK-switch based on regrasping was proposed to bridge the gap between different components of the configuration space that would otherwise be disconnected.
	The authors further proposed a certified-complete planner of the bimanual system in \cite{lertkultanon2018certified}, which could be applied to bring an object from the initial stable placement toward a goal stable placement.
	Likewise, configuration disconnectivity was also discussed in \cite{jang2022motion} and \cite{wu2022global}, where the manipulators completed the manipulation through regrasping in the experiments.
	
	It is noted that the above methods always rely on regrasping, which may impose further difficulties.
	First, regrasping leads to inefficient manipulation.
	The manipulator needs to release the object, adjust to the other configuration, and regrasp the object.
	During the process, other manipulators and the object usually remain stationary \cite{ xian2017closed, lertkultanon2018certified, jang2022motion}, which is time-consuming.
	Second, feasibility issues under heavy loads.
	In practical applications, the rest of manipulators may not be able to hold the object during the regrasping process.
	Furthermore, regrasping may cause vibration and introduce impact issues.
	These problems need additional solutions, such as leveraging environmental support \cite{lertkultanon2017planning}, which also results in inefficiency.
	
	Fortunately, with the additional degrees of freedom provided by the mobile platform, MMMS give a chance to accomplish complex flipping manipulations without regrasping.
	However, the redundancy brought by the mobile platform of MMMS also presents another challenge, as the motion planning is extended to a higher-dimensional space.
	Existing research on motion planners for mobile manipulators can be broadly categorized into two categories.
	The first category is decoupled planning \cite{brock2001decomposition}, where the planning is divided into two steps.
	The first step is path planning of the mobile platform by graph search algorithms and the second step is the planning of the manipulators. For example, door traversal of mobile manipulators was considered in \cite{jang2023motion}.
	Similarly, the method can be found in \cite{thakar2020manipulator}.
	Decoupled planning is relatively simple and easy to implement, but it cannot effectively utilize the mobility of the mobile platform and the manipulability of the manipulators.
	The second category is coordinated planning, which moves the mobile platform and manipulator simultaneously.
	In such cases, there are infinite configurations that can be planned for the object manipulation due to redundancy, and it can be challenging to determine a specific configuration\cite{huang2000coordinated}.
	For this reason, many sample-based or optimization-based methods were proposed for redundancy planning \cite{berenson2011task}\cite{oriolo2005motion}.
	However, given a complex manipulation, like flipping motion, there is still no answer whether there is a planning result without regrasping or with a minimal number of regrasping.
	
	\subsection{Contribution}
	To enable complex object manipulation for MMMS, this paper proposes a novel planning framework.
	Given the desired object trajectory and stable grasping pose set, the planning framework includes coordinated platform planning and regrasping planning, both of which can address the configuration disconnectivity problem of the manipulator.
	In the planning process, the whole object trajectory can be divided into different segments based on the given grasping pose set.
	These trajectory segments can be further optimally assigned to different manipulators, which is formulated into a set covering problem.
	By analyzing different coverage combinations, we can quickly determine whether the manipulation can complete the task without regrasping or with the minimal number of regrasping.
	After this, motions along assigned trajectory segments can be planned through existing methods.
	
	
	The main contributions of this paper are threefold.
	\begin{itemize}[leftmargin = *]
		\item[1)]
		A novel planning framework is proposed for complex flipping manipulation of MMMS.
		Coordinated platform motions and regrasping motions are enrolled to address the configuration disconnectivity problem.
		The planner can quickly determine whether a complex manipulation task can be completed without regrasping or not.
		\item[2)]
		The planning is further formulated as a set cover problem based on the coverable trajectory segment, where priority is given to avoid regrasping to ensure a minimal number of regrasping.
		\item[3)]
		The proposed planning framework has been validated by extensive physical experiments.
		Compared to planners that focus on transportation tasks \cite{zhang2021task,jiao2015transportation,alonso2017multi}, the proposed planning method adapts to a broader range of complex manipulations.
	\end{itemize}

	The remainder of this paper is organized as follows.
	In Section II, multiple mobile manipulator systems are introduced and the planning problem is formulated.
	In Section III, the proposed motion planning framework is presented in detail.
	In Section IV, both simulation and experimental results are given.
	Finally, Section V concludes the paper and outlines future work.

	\section{Problem Formulation}

	\subsection{Preliminaries}
	As shown in Fig.~\ref{MMMS}, it illustrates multiple mobile manipulators manipulating an object.
	Let $\mathbb{W}$ and $\mathbb{O}$ be the frames of the world and object, respectively.
	Let $\mathbb{P}_i$, $\mathbb{B}_i$, and $\mathbb{E}_i$ be the frames of the mobile platform, the base of the manipulator, and the end-effector of the $i$-th mobile manipulator, respectively, where $i=1,\ldots ,n$ and $n$ is the number of mobile manipulators.
	${\bf{T}} \in SE(3)$ represents the pose of a frame with respect to another frame.
	For example,  $^{\mathbb{W}}{\bf{T}}_{\mathbb{O}}$ represents the pose of frame $\mathbb{O}$ in frame $\mathbb{W}$.
	Based on frame transformations, the following fundamental relationships can be obtained
	\begin{equation}
		\label{transformations}
		^{\mathbb{W}}{\bf{T}}_{\mathbb{O}} = {^{\mathbb{W}}{\bf{T}}_{\mathbb{P}_i}} {^{\mathbb{P}_i}{\bf{T}}_{\mathbb{B}_i}}{^{\mathbb{B}_i}{\bf{T}}_{\mathbb{E}_i}}{^{\mathbb{E}_i}{\bf{T}}_{\mathbb{O}}}
	\end{equation}
	which can be regarded as a chain in the system.
	It should be noted that ${^{\mathbb{P}_i}{\bf{T}}_{\mathbb{B}_i}}$ is invariant and solely depends on the robot itself, specifically, the location of the manipulator mounted on the mobile platform.
	
	The configurations of the $i$-th mobile manipulator is denoted as $\bm{q}_i = (\bm{q}^{\rm T}_{p,i}, \bm{q}^{\rm T}_{m,i})^{\rm T}$, where $\bm{q}_{p,i} \in \mathbb{R}^3$ refers to the configuration of the mobile platform, and $\bm{q}_{m,i} \in \mathbb{R}^6$ refers to the configuration of the manipulator.
	Let $\mathcal{C}^i$ denote the whole configuration space of the $i$-th robot and $\mathcal{C}^i \subset \mathbb{R}^9$.
	Due to self-collisions and collisions between the robot and the object, not all configurations are valid in $\mathcal{C}^i$.
	The obstacle-free configuration space is denoted as $\mathcal{C}^{i}_{\textit{free}}$.
	
	\subsection{Problem Definition}
	
	Let $\gamma_{\textit{obj}} = [0, 1]$ denote the whole trajectory of the object.
	The trajectory of the object is given as $\tau_{\textit{obj}}: [0, 1] \to [{^{\mathbb{W}}{\bf{T}}_{\mathbb{O}}}(0), {^{\mathbb{W}}{\bf{T}}_{\mathbb{O}}}(1)]$.
	Then, the trajectory of the end-effector of the $i$-th robot can be obtained by $\tau_{\textit{i}} = \tau_{\textit{obj}}{^{\mathbb{O}}{\bf{T}}_{\mathbb{E}_i}}$, and ${^{\mathbb{O}}{\bf{T}}_{\mathbb{E}_i}}$ is determined by the specific grasping pose.
	
	Let $g$ denote a grasping pose of the manipulator to the object.
	Since stable grasping pose planning is not the focus of this paper, although it is itself a challenging problem, it is assumed that the grasping pose set is given in advance.
	All possible grasping poses are denoted as $\mathcal{G} = \left\{g_{1}, \dots, g_{m}\right\}$.
	$m$ is the number of given grasping poses, where $m$ must be greater than or equal to $n$.
	Therefore, the planning problem can be defined as
	\begin{equation}
		\label{problem}
		\begin{aligned} 
			\text{Input:}& \ \  \tau_{\textit{obj}}: [0, 1] \to [{^{\mathbb{W}}{\bf{T}}_{\mathbb{O}}}(0), {^{\mathbb{W}}{\bf{T}}_{\mathbb{O}}}(1)], \  \mathcal{G}\\
			\text{Output:}& \ \ \tau_{i}: [0, 1] \to \begin{cases} [{^{\mathbb{W}}{\bf{T}}_{\mathbb{P}_i}}(0), {^{\mathbb{W}}{\bf{T}}_{\mathbb{P}_i}}(1)] \\
				[{^{\mathbb{B}_i}{\bf{T}}_{\mathbb{E}_i}}(0), {^{\mathbb{B}_i}{\bf{T}}_{\mathbb{E}_i}}(1)] 
			\end{cases} \\
			\text{s.t}& \ \ \text{Eq}.\eqref{transformations}
		\end{aligned}
	\end{equation}
	
	The kinematics transformations of the mobile platform and the manipulator are defined as $f_{p}$ and $f_{m}$, respectively. Then we can obtain $f_{p}(\bm{q}_{i,p}) = {^{\mathbb{W}}{\bf{T}}_{\mathbb{P}_i}}$.
	Similarly, $f_{m}(\bm{q}_{i,m}) = {^{\mathbb{B}_i}{\bf{T}}_{\mathbb{E}_i}}$.
	Since ${^{\mathbb{P}_i}{\bf{T}}_{\mathbb{B}_i}}$ is only related to the mounting pose of manipulators on the mobile platform, the kinematics transformations of the mobile manipulator can be defined as follows.
	\begin{equation}
		\label{kinematics}
		f(\bm{q}_i) {^{\mathbb{E}_i}{\bf{T}}_{\mathbb{O}}} = {^{\mathbb{W}}{\bf{T}}_{\mathbb{O}}}, \ \bm{q}_i \in \mathbb{R}^9
	\end{equation}
	Obviously, Eq.\eqref{kinematics} is a redundant motion planning for the mobile manipulator.
	According to Eq.\eqref{transformations}, it can be derived that
	\begin{equation}
		\label{closedchain}
		f^{-1}(\bm{q}_j)f(\bm{q}_i) = {^{\mathbb{E}_j}{\bf{T}}_{\mathbb{E}_i}}, \ i \neq j
	\end{equation}
	Eq.\eqref{closedchain} can be regarded as the closed-chain constraints of the MMMS, indicating that end-effectors of the manipulators must maintain a specific relative pose.
	If no regrasping occurs during the manipulation, ${^{\mathbb{E}_j}{\bf{T}}_{\mathbb{E}_i}}$ should remain unchanged.

	\begin{figure*}[htbp]
		\includegraphics[width=\textwidth]{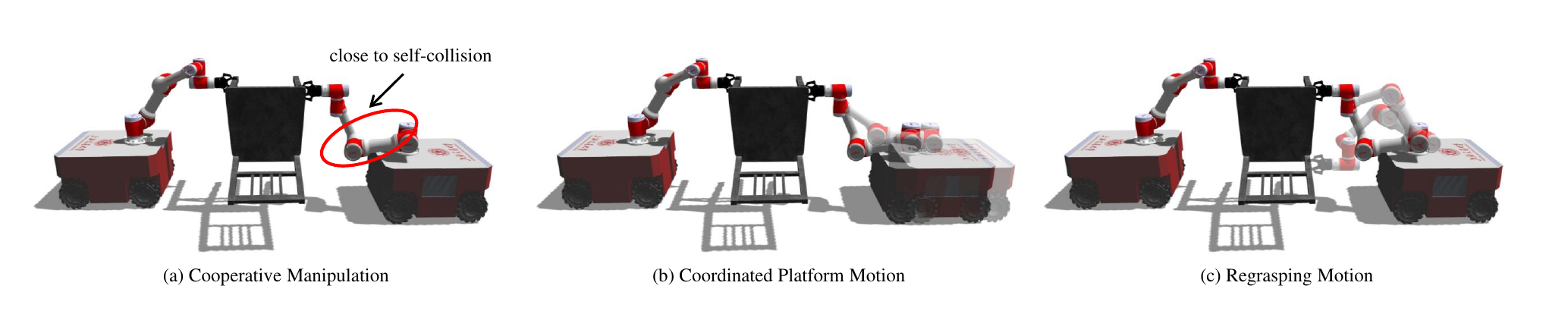}
		\centering
		\caption{
			Different motions of mobile manipulators enhance connectivity when manipulating objects.
			(a) Due to self-collision or joint angle limits of the robots, the expected trajectory of the object may not be completed.
			(b) Moving the platform to increase connectivity while keeping the grasping pose unchanged.
			(c) Regrasping transitioning to other configurations to enhance connectivity.
		}
		\label{DifferentPath}       
	\end{figure*}

	Substituting Eq.\eqref{kinematics}-\eqref{closedchain} into the problem given in \eqref{problem}, the planning problem can be transformed into the configuration space of the mobile manipulator
	\begin{equation}
		\label{problem_new}
		\begin{aligned} 
			\text{Input}& \ \  \tau_{\textit{obj}}: [0, 1] \to [{^{\mathbb{W}}{\bf{T}}_{\mathbb{O}}}(0), {^{\mathbb{W}}{\bf{T}}_{\mathbb{O}}}(1)], \  \mathcal{G} \\
			\text{Output}& \ \ \tau_{i}: [0, 1] \to [\bm{q}_i(0), \bm{q}_i(1)]  \\
			\text{s.t}& \ \ \begin{cases}  f(\bm{q}_{i}(t)) = {^{\mathbb{W}}{\bf{T}}_{\mathbb{O}}(t)}{^{\mathbb{O}}{\bf{T}}_{\mathbb{E}_i}(t)}  \\
				\bm{q}_{i}(t) \in \mathcal{C}^{i}_{\textit{free}} \ t \in [0,1]
			\end{cases} \\
		\end{aligned}
	\end{equation}
	As clarified before, some research has employed the regrasping approach.
	If $\bm{q}_{m,i}(t_{k})$ approaches the boundary of ${C}^{i}_{\textit{free}}$.
	The robot can regrasp to change the grasping pose from $g$ to $g'$, and correspondingly, the configuration is transformed to $\bm{q}'_{m,i}(t_{k})$.
	$\bm{q}'_{m,i}(t_{k})$ may have a larger range away from the boundaries.
	Thus, the connectivity of the configuration space under closed-chain constraints can be improved.
	Due to the additional degrees of freedom provided by the mobile platform, the manipulator can adjust its configuration without releasing the object, thus can also avoid regrasping.
	Both approaches are incorporated into our planning framework.

	\section{The Proposed Motion Planner}
	
	This paper focuses on motion planning for robots, assuming that the trajectory of the object and grasping poses are already provided.
	Two typical motions, coordinated platform motion and regrasping motion, are introduced, both of which enhance the configuration connectivity of the manipulator.
	The planning processes for the two corresponding motions are discussed.
	Then, a comprehensive framework is designed, which prioritizes coordinated platform motions to achieve manipulation without regrasping.
	If the task cannot be completed, then it considers regrasping with the minimal number.

	\subsection{Strategy Classification}

	Based on the movement of mobile manipulators during object manipulation within a trajectory segment $\gamma = [\alpha_1, \alpha_2]$, the following types of motions are categorized:
	\begin{itemize}[leftmargin = *]
		\item[$\bullet$]
		\textit{Coordinated Platform}:  ${^{\mathbb{E}_i}{\bf{T}}_{\mathbb{O}}}(t)$ = ${^{\mathbb{E}_i}{\bf{T}}_{\mathbb{O}}}(\alpha_1)$, $t \in [\alpha_1,\alpha_2]$.
		\item[$\bullet$]
		\textit{Regrasping}:  $\tau_{B}: [\alpha_1, \alpha_2] \to [{^{\mathbb{E}_i}{\bf{T}}_{\mathbb{O}}}(\alpha_1), {^{\mathbb{E}_i}{\bf{T}}_{\mathbb{O}}}(\alpha_2)]$.
	\end{itemize} 
	Two different motions are shown in Fig.~\ref{DifferentPath}, both of which can enhance the configuration connectivity of the manipulator by switching configuration.
	In Fig.~\ref{DifferentPath}(b), the platform collaborates with the manipulator to avoid configuration disconnectivity.
	Fig.~\ref{DifferentPath}(c) represents a regrasping motion, where the manipulator releases the object, shifts to a desired configuration, and subsequently grasps the object.
	During regrasping, there will be a change in the closed-chain constraints.
	It is assumed that during the regrasping process, other robots are kept stationary and can hold the object stable.
	The assumption is common for safety manipulation \cite{xian2017closed}\cite{jang2022motion}.

	\subsection{Coordinated Platform Planning}
	
	\begin{algorithm}[t]
		\caption{Coordinated Platform Planner}
		\label{CoordinatedPlatformPlanner}
		\KwIn{$\tau_{\textit{obj}}(\gamma): [\alpha_1, \alpha_2] \to [{^{\mathbb{W}}{\bf{T}}_{\mathbb{O}}}(\alpha_1), {^{\mathbb{W}}{\bf{T}}_{\mathbb{O}}}(\alpha_2)], g$.}
		\KwOut{$\tau_{A}(\gamma): [\alpha_1, \alpha_2] \to [\bm{q}(\alpha_1), \bm{q}(\alpha_2)]$.}
		$\tau_{A}, \tau$ $\gets \varnothing$; $\gamma_1 \gets \varnothing$; \\
		\While{\rm $\gamma_1 \prec \gamma$}
		{
			$\tau, \gamma_1$ $\gets$ \texttt{PlanManipulator}$(\tau_{\textit{obj}}(\gamma \setminus \gamma_1), g)$; \\
			$\tau_{A}$ $\gets$ \texttt{TrajectorySynthesis}$(\tau_{A}, \tau)$; \\
			\If{\rm $\gamma_1 == \gamma$}
			{
				\textbf{break};
			}
			$\tau, \gamma_1$ $\gets$ \texttt{ConnetPlan}$(\tau_{\textit{obj}}(\gamma \setminus \gamma_1), g)$; \\
			$\tau_{A}$ $\gets$ \texttt{TrajectorySynthesis}$(\tau_{A}, \tau)$; \\
		}
		\Return $\tau_{A}$;
	\end{algorithm}
	
	Coordinated platform planning requires planning a collision-free configuration trajectory for the robot in a trajectory segment $\gamma = [\alpha_1, \alpha_2]$.
	Specifically, when the grasping pose $g$ remains unchanged, plan a configuration trajectory $\tau_{A}(\gamma): [\alpha_1, \alpha_2] \to [\bm{q}(\alpha_1), \bm{q}(\alpha_2)]$ for the robot.
	It should be pointed out that the manipulator is generally more accurate and flexible than the mobile platform, and therefore completing the manipulation task primarily relies on the motion of the manipulator.
	Hence, the platform will remain as stationary as possible.
	With this motivation, the planning process is detailed in Algorithm \ref{CoordinatedPlatformPlanner} and some key functions are explained as follows.
	
	\begin{itemize}[leftmargin = *]
		\item[$\bullet$]
		The input is the given object trajectory $\tau_{\textit{obj}}(\gamma)$ with trajectory segment $\gamma$.
		Let $\gamma_{1} \in \gamma$ denote the trajectory segment that has been planned, and $\gamma \setminus \gamma_{1}$ denote the trajectory segment that has not been planned.
		Therefore, at the beginning, $\gamma_{1} = \varnothing$ and $\gamma \setminus \gamma_{1} = \gamma$, as shown in the Line 1.
		The loop in Algorithm 1 plans configurations for the robot until the whole trajectory segment $\gamma$ is planned.
		\item[$\bullet$]
		\texttt{PlanManipulator} plans the trajectory of manipulators while keeping the mobile platform stationary.
		Most likely, relying solely on the manipulator cannot finish the manipulation due to limited reachable space or self-collisions.
		In this case, at the end of the planned configuration trajectory, the manipulator may approach the boundary of the valid configuration space, which requires subsequent process.
		\item[$\bullet$]
		\texttt{TrajectorySynthesis} integrates the planned trajectories.
		\item[$\bullet$]
		\texttt{ConnectPlan} plans the collaborative configuration trajectory of the platform and manipulator, enhancing the connectivity of the manipulator.
		In order to avoid the curse of dimensionality, this paper adopts a decoupled planning approach similar to \cite{jang2023motion}\cite{thakar2020manipulator}.
		Assuming the planned trajectory is $\tau(\gamma_1): [\alpha_1, \beta_1] \to [\bm{q}(\alpha_1), \bm{q}(\beta_1)]$, $\bm{q}(\beta_1) = (\bm{q}^{\rm T}_{p}(\beta_1), \bm{q}^{\rm T}_{m}(\beta_1))^{\rm T}$, and the remaining trajectory segment is $\gamma_2 = [\beta_1, \alpha_2]$.
		The desired configuration of the platform is determined through the following optimal planning
		\begin{equation}
			\label{mmplanning}
			\begin{aligned} 
				\bm{q}_{p}& =\ \text{argmax} \ \beta_2' \\
				\text{s.t} \ \ &  \beta_1-\beta_1' \ge \xi	\\
				& \bm{q}_{m}(\beta_1) \in [\bm{q}_{m}(\beta_1'), \bm{q}_{m}(\beta_2')] 
			\end{aligned}
		\end{equation}
		where $[\beta_1', \beta_2'] \gets \texttt{PlanManipulator}(\tau_{\textit{obj}},\bm{q}_{p}, g)$.
		$\xi$ is a positive coefficient, where a larger value indicates more shared solutions between the two trajectory segments $[\beta_1', \beta_2']$ and $[\alpha_1, \beta_1]$.
		After obtaining $\bm{q}_{p}$, the direction of the platform's movement is determined, and then $\tau$ can be planned.
		This optimization is aimed at ensuring the connectivity of the manipulator when platform moves and enhancing subsequent planning if necessary.
	\end{itemize} 
	
	In the planning phase, the motions of the platform are aimed at facilitating the connection of configurations that would otherwise be disconnected of the manipulator.
	Hence, the platform moves only when the manipulator is close to the configuration boundary, which will be further demostrated in the experiment section.
	
	\subsection{Regrasping Planning}
	
	\begin{algorithm}[t]
		\caption{Regrasping Planner}
		\label{RegraspingPlanner}
		\KwIn{$\tau(\gamma): [\alpha_1,\alpha_2] \to [\bm{q}(\alpha_1),\bm{q}(\alpha_2)], g, g'$.}
		\KwOut{$\tau_{B}: [\alpha_2, \alpha_3] \to [\bm{q}(\alpha_2), \bm{q}(\alpha_3)]$.}
		$\left\{\bm{q}\right\}$ $\gets$ \texttt{IKsolution}$(g)$; \\
		$\bm{q}(\alpha_3)$ $\gets$ \texttt{ManiOptimization}$(\left\{\bm{q}\right\})$; \\
		$\tau_{B}$ $\gets$ \texttt{Regrasping}$(\bm{q}(\alpha_2), \bm{q}(\alpha_3))$; \\
		\Return $\tau_{B}$;
	\end{algorithm}
	
	Under closed-chain constraints, configurations may need regrasping to achieve connectivity.
	For a robot, it is assumed that the trajectory before regrasping has already been planned and is denoted as $\tau(\gamma): [\alpha_1,\alpha_2] \to [\bm{q}(\alpha_1),\bm{q}(\alpha_2)]$.
	$\bm{q}(\alpha_2)$ suffers the configuration disconnectivity and thus the grasping pose needs to switch from $g$ to $g'$.
	Then, the regrasping planning problem is to plan $\tau_{B}: [\alpha_2, \alpha_3] \to [\bm{q}(\alpha_2),\bm{q}(\alpha_3)]$, where the new closed-chain constraint ${^{\mathbb{E}}{\bf{T}}_{\mathbb{O}}}$ must be re-established at $\bm{q}(\alpha_3)$.
	Hence, the most critical part is to determine $\bm{q}(\alpha_3)$.
	
	Here, we consider the following optimal planning
	\begin{equation}
		\label{regrasping}
		\begin{aligned} 
			\bm{q}(\alpha_3) =& \ \text{argmax} \ det\sqrt{\bm{J}_{m}(\bm{q}(\alpha_3))\bm{J}_{m}^{\rm T}(\bm{q}(\alpha_3))} \\
			\text{s.t} &\ \ f(\bm{q}(\alpha_3)) {^{\mathbb{E}}{\bf{T}}_{\mathbb{O}}} = {^{\mathbb{W}}{\bf{T}}_{\mathbb{O}}} \\
		\end{aligned}
	\end{equation}
	where $\bm{J}_{m}$ is the Jacobian matrix of the manipulator, and $det\sqrt{\bm{J}_{m}(\bm{q}(\alpha_3))\bm{J}_{m}^{\rm T}(\bm{q}(\alpha_3))}$ is the manipulability of robots \cite{yoshikawa1985manipulability}.
	This optimization aims to enhance the robot's operational capabilities after regrasping and facilitate the subsequent planning.
	The planning process is detailed in Algorithm \ref{RegraspingPlanner} and some key functions are explained as follows.
	\begin{itemize}[leftmargin = *]
		\item[$\bullet$]
		\texttt{IKsolution} and \texttt{ManiOptimization} are employed to solve the optimization problem. 
		The configurations of mobile platform are sampled first, and then the configurations of manipulator can be solved through the inverse kinematics solvers.
		Finally, the optimal configuration with highest manipulability is selected for regrasping.
		\item[$\bullet$]
		\texttt{Regrasping} plans a trajectory for robot from the initial configuration to the planned configuration.
		After releasing the object, the robot can move relatively freely and the trajectory can be planned easily.
		The only consideration during \texttt{Regrasping} is collision avoidance.
		In the trajectory segment $[\alpha_2, \alpha_3]$, the robot regrasps the object and other robots remain stationary.
	\end{itemize}

	\subsection{General Framework of Planner}

	\begin{figure}[t]
		\includegraphics[width=\columnwidth]{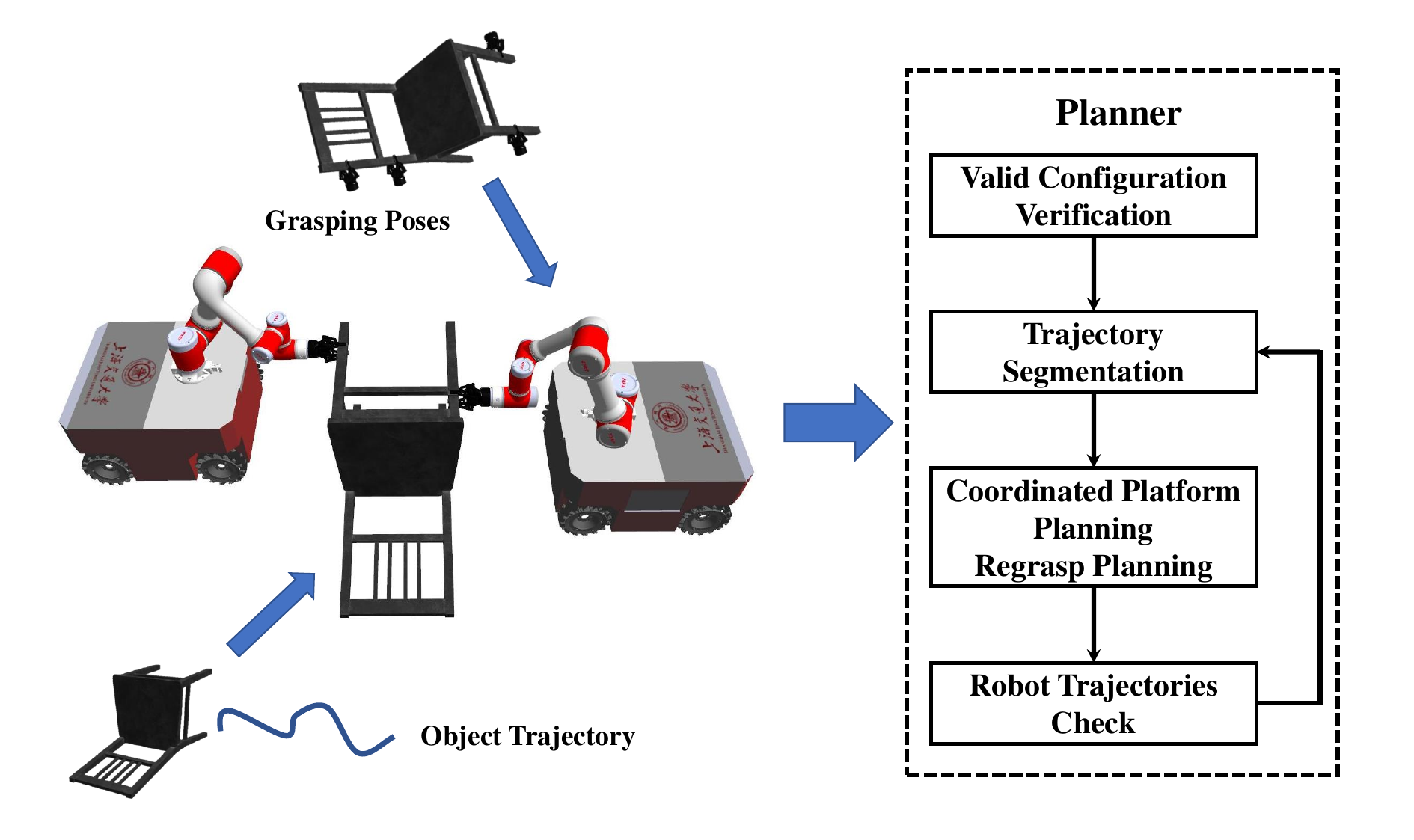}
		\centering
		\caption{
			The framework of our planner.
			The object trajectory and grasp poses serve as inputs to the planner.
		}
		\label{Planner}       
	\end{figure}

	\begin{algorithm}[t]
		\caption{Global Planner}
		\label{globalplanner}
		\KwIn{$\tau_{\textit{obj}}(\gamma_{\textit{obj}})$, $\mathcal{G} = \left\{ g_{1}, \dots, g_{m} \right\}$.}
		\KwOut{$\tau_{i}: [0, 1] \to [\bm{q}_i(0), \bm{q}_i(1)], i = 1,\dots,n$.}
		$\Gamma \gets$ \texttt{IKcheck}$(\tau_{\textit{obj}},  \mathcal{G})$; \\
		\If{ $\Gamma \prec \gamma_{\textit{obj}}$ }
		{
			\Return $\varnothing$;
		}
		\While{\rm \texttt{TrajectoryCheck}$(\tau_{i})$ is true}
		{
			$\mathcal{S}_1, \dots, \mathcal{S}_n$ $\gets$ \texttt{Assignment}$(\Gamma)$;  \\
			\For{$i=1$ \rm{to} $n$}
			{
				$\tau_{i}$ $\gets$ \texttt{Coor\_Platform}$(\mathcal{S}_i)$; \\
				$\tau_{i}$ $\gets$ \texttt{Regrasping}$(\mathcal{S}_i)$; \\
			}
		}
		
		\Return $\tau_{1}, \dots, \tau_{n}$;
	\end{algorithm}
	
	\begin{figure}[t]
		\includegraphics[width=\columnwidth]{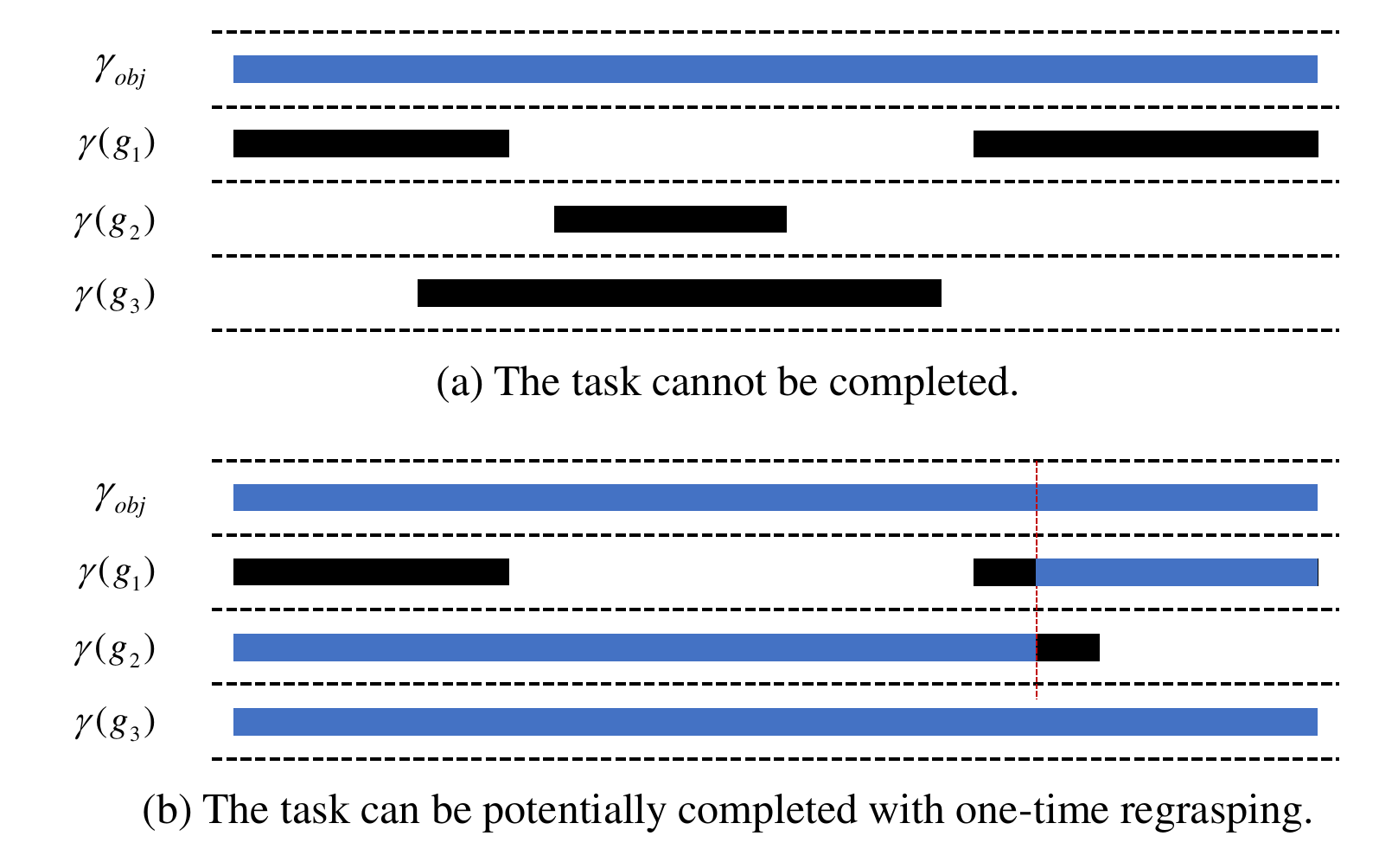}
		\centering
		\caption{
			The whole trajectory is divided into different segments with different grasping poses for the robots based on minimizing the number of regrasping.
			An example of two robots.
		}
		\label{Division}       
	\end{figure}
	
	The framework of our planner is illustrated in Fig.~\ref{Planner}.
	Object trajectory $\tau_{\textit{obj}}(\gamma_{\textit{obj}})$ and grasping poses $\mathcal{G}$ are the inputs.
	For all possible grasping poses of the $i$-th robot, valid configurations are verified.
	Based on the verification, the trajectory segments that each grasping pose covers can be roughly obtained.
	Then, the whole trajectory segment can be planned or be divided into different segments, which minimizes the regrasping number.
	Finally, specific planning is conducted for each trajectory segment.
	The planning process is detailed in Algorithm \ref{globalplanner} and some key functions are explained as follows.
	
	\begin{figure*}[htbp]
		\includegraphics[width=\textwidth]{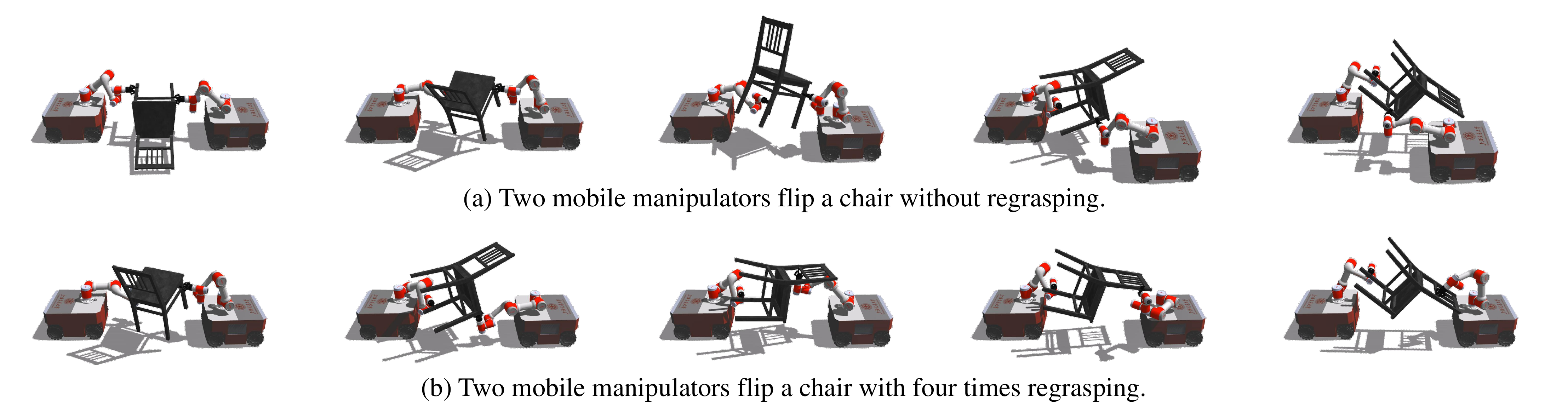}
		\centering
		\caption{
			Flipping manipulation of a chair by two mobile JAKA Zu7 manipulators.
			Comparisons between the proposed planner and other methods.
		}
		\label{simulation1}       
	\end{figure*}
	
	\begin{figure*}[htbp]
		\includegraphics[width=\textwidth]{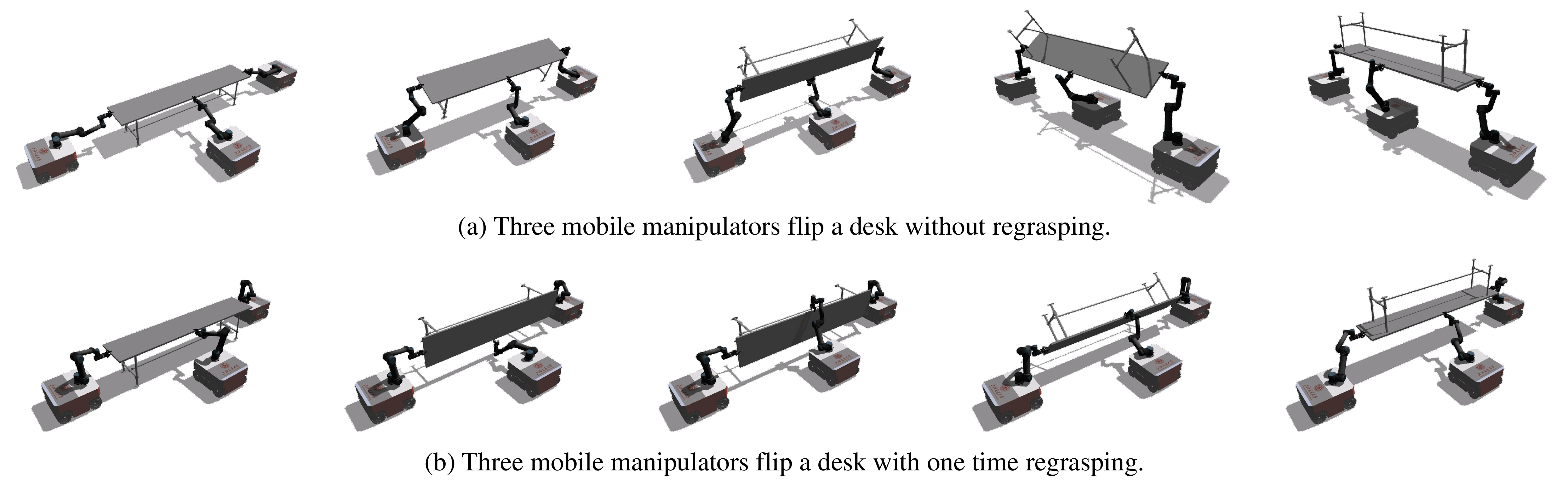}
		\centering
		\caption{
			Flipping manipulation of a desk by three mobile UR5 manipulators.
			Trajectories of the desk are different to test the proposed planner.
		}
		\label{simulation2}       
	\end{figure*}
	
	\begin{itemize}[leftmargin = *]
		\item[$\bullet$]
		\texttt{IKcheck} verifies whether feasible configurations exist under the given trajectory and closed-chain constraints.
		Specifically, for the all possible grasping poses $\mathcal{G} = \left\{g_{1}, \dots, g_{m}\right\}$, it checks for the existence of $\bm{q}_i$ that satisfies Eq.\eqref{kinematics}.
		Let $\gamma(g_j) \subset \gamma_{\textit{obj}}, j = 1,...,m$ denote the trajectory segment with feasible configurations under grasping pose $g_{j}$.
		The output of \texttt{IKcheck} is $\Gamma = \gamma(g_1)\cup \dots \cup \gamma(g_m) $.
		If $\Gamma \prec \gamma_{\textit{obj}}$, it signifies that there are no feasible configurations in certain points regardless of the grasping poses employed.
		A case is shown in Fig.~\ref{Division}(a).
		In this case, the planning is bound to fail and the remaining steps will be skipped.
		\item[$\bullet$]
		\texttt{Assignment} chooses grasping poses and their corresponding trajectory segments $\gamma(g_j)$, then allocates them to robots.
		It is executed based on minimizing the number of regrasping.
		The problem can be formulated as
		\begin{equation}
			\label{divide}
			\begin{aligned} 
				\underset{\gamma(g_j)}{\text{min}} \ k & \\
				\text{s.t} \ \ \gamma(g_1) \cup  \dots \cup & \gamma(g_k) = \gamma_{\textit{obj}} \\
				\gamma(g_j) \subset  \Gamma, & \ j \in [1,k]
			\end{aligned}
		\end{equation}
		Eq.(\ref{divide}) is a set covering problem, and all combination schemes $\mathcal{S}$ can be found\cite{beasley1996genetic}.
		The minimum number of regrasping corresponds to the minimum set cover.
		Then, the schemes are allocated to the robots in ascending order.
		Since the same grasping pose can only be used by one robot at the same time, the following constraint must also be satisfied
		\begin{equation}
			\label{constraints}
			\begin{aligned} 
				\forall \ t \in \gamma_{\textit{obj}}, \ \mathcal{S}_1(t) \cup  \dots \cup \mathcal{S}_n(t) = \varnothing
			\end{aligned}
		\end{equation}
		A case is shown in Fig.~\ref{Division}(b).
		The covering scheme $\mathcal{S}_1 = \left\{ \gamma(g_2), \gamma(g_1) \right\}$ and $\mathcal{S}_2 = \left\{ \gamma(g_3) \right\}$ are chosen for robots.
		\item[$\bullet$]
		\texttt{Coor\_Platform} and \texttt{Regrasping} are carried out through Algorithm \ref{CoordinatedPlatformPlanner} and Algorithm \ref{RegraspingPlanner}, respectively.
		\item[$\bullet$]
		\texttt{TrajectoryCheck} is employed to verify whether the generated trajectory $\tau_{i}$ meets the requirements and whether self-collisions occur during execution.
	\end{itemize}

	After planning the configuration trajectories of all robots, we can carry out time synchronization and motion planning on them to complete the planning task.

	\section{Simulation and Experiment}

	\subsection{Simulation and Experimental Setup}
	
	The mobile platforms are omnidirectional wheeled and manipulators are UR5 and JAKA Zu7, and the end of each manipulator is equipped with a Robotiq 2F-85 gripper.
	All our simulations are performed on a computer with Intel Core i7-10700 CPU at 2.90 GHz (16 cores) and 64GB RAM.
	The results are obtained by the simulation platform ROS-Gazebo based on C++.
	Real-world experiments are conducted on the mobile JAKA Zu7 robots.
	Each robot is controlled separately by its onboard computer with Intel Core i7-1165G7 CPU at 2.80 GHz (4 cores) and 16GB RAM.
	
	In physical experiments, it is almost impossible for robots to achieve consistency as planned due to many unknown disturbances.
	Due to the tight connection between the gripper and the object, inconsistency may lead to excessive interaction forces, resulting in potential damage\cite{ren2020fully}.
	The leader-follower structure is enrolled to partly address the problem and ensure safety.
	Assuming the leader robot of the system is $l$.
	When executing the planned motions, the configurations at time $t$ should satisfy the closed-chain constraints
	\begin{equation}
		\label{closedchain_planned}
		\begin{aligned} 
			f^{-1}(\bm{q}_l(t)) f_p(\bm{q}_{i,p}&(t)) {^{\mathbb{P}_i}{\bf{T}}_{\mathbb{B}_i}} f_m(\bm{q}_{i,m}(t)) = {^{\mathbb{E}_l}{\bf{T}}_{\mathbb{E}_i}}(t) \\
			& i = 1, \dots, n \ i \neq l
		\end{aligned}
	\end{equation}
	In order to maintain Eq.\eqref{closedchain_planned}, the planned configuration of the follower robot $\bm{q}_i(t)$ needs to be instantaneously modified to  $\bm{q}'_i(t)$ according to the actual configuration of the leader.

	\subsection{Simulation and Experimental Results}
	
	\begin{figure*}[htbp]
		\includegraphics[width=\textwidth]{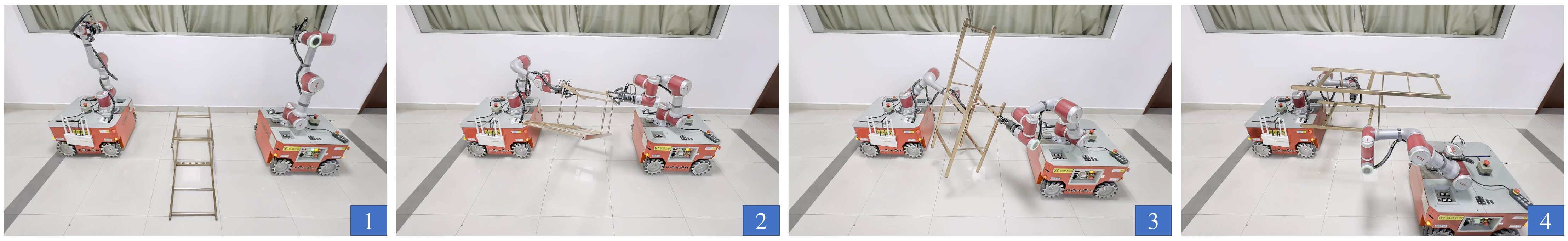}
		\centering
		\caption{
			Flipping manipulation of a chair by two mobile JAKA Zu7 manipulators in real world.
			The setup is similar to Fig.~\ref{simulation1}(a).
		}
		\label{experiment}       
	\end{figure*}
	
	\begin{figure*}[htbp]
		\includegraphics[width=\textwidth]{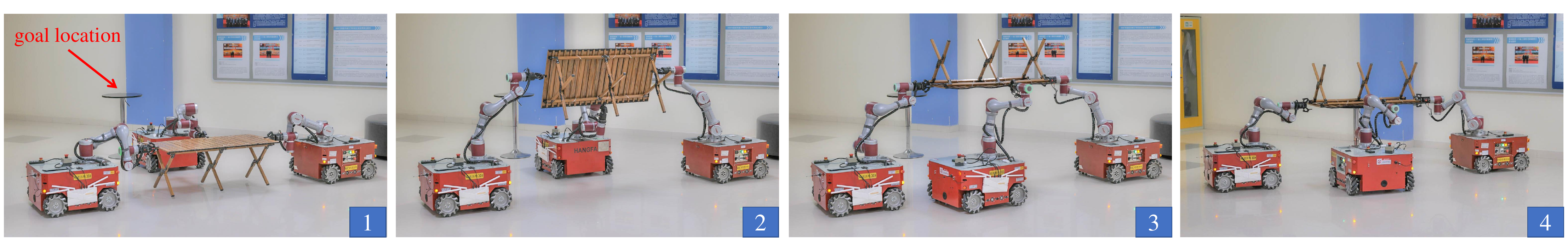}
		\centering
		\caption{
			A manipulation task requires placing the table in a specified location.
			Due to the limited reachable space of the robots, three mobile JAKA Zu7 manipulators have to flip the object by 180 degrees before being able to place it.
		}
		\label{application}       
	\end{figure*}
	
	\subsubsection{Flip a Chair in Simulation}
	
	The first set of simulations uses two mobile JAKA Zu7 manipulators to flip a chair.
	During the flipping, the position of the chair remains unchanged, but the orientation needs to change from (90, 0, 0) degrees to (-180, 45, 90) degrees.
	
	The first time, the motions of robots are generated through the proposed planner.
	As shown in Fig.~\ref{simulation1}(a), the robots successfully complete the manipulation without the need for regrasping.
	The robot on the right side needs to move its platform during the manipulation.
	The manipulation takes 18 seconds.
	The second time, the method of regrasping is tried \cite{zhang2021task,jiao2015transportation,alonso2017multi}, and the results are shown in Fig.~\ref{simulation1}(b).
	The robots also successfully complete the manipulation.
	The robot on the right side regrasps the chair and changes the grasping pose four times to achieve the desired manipulation.
	During the regrasping, the robot on the left side remains stationary.
	The manipulation takes 36 seconds.
	These comparisons show the second contribution of this paper.

	\subsubsection{Flip a Desk in Simulation}
	
	The second set of simulations uses three mobile UR5 manipulators to flip a desk.
	The orientation of the desk needs to change from (0, 0, 0) degrees to (180, 0, 0) degrees.
	The motions are generated through the proposed planner.
	
	The first time, the trajectory of the desk is higher, allowing the robot to pass through below.
	The robots achieve the manipulation task without the need for regrasping, as shown in Fig.~\ref{simulation1}(a).
	The second time, the trajectory of the desk is lower.
	There is not enough space for the robot to pass through.
	As shown in Fig.~\ref{simulation1}(b), the robot is close to collision.
	In order to continue manipulation, the robot has to regrasp to change the grasping pose.
	When regrasping, the robot selects the upper part of the desk because it could complete the subsequent trajectory in the planning stage, thus requiring only one regrasping.
	
	In such complex manipulation, the manipulators may easily suffer configuration disconnectivity due to limited configuration space or collisions.
	However, during regrasping, the rest robots need to remain stationary while and wait, which greatly reduces the efficiency.
	The manipulation in Fig.~\ref{simulation1}(a) only takes 15 seconds, while in Fig.~\ref{simulation1}(b), it takes 30 seconds.
	Besides, in reality, the robot on the left may not be able to support the object independently.
	Therefore, in the proposed planner, coordinated platform motions are preferred.
	If the manipulation cannot be achieved, as shown in Fig.~\ref{simulation1}(b), regrasping is considered with the minimum number.
	These comparisons show the first contribution of this paper.

	\subsubsection{Physical Experiments}
	
	Physical experiments similar to Fig.~\ref{simulation1} are conducted, and the results are shown in Fig.~\ref{experiment}.
	The proposed method successfully plans coordinated motions of the platform to enhance connectivity for the manipulator.

	Finally, we conduct a comprehensive experiment requiring a table to be transported and placed in a specified location.
	The results are shown in Fig.~\ref{application}, where the first image highlights the desired placement location.
	If a conventional planner is employed, this task would be impossible to complete since the table would need to be lifted to a height beyond the reachable range of the robot.
	However, within our planning framework, the table can be manipulated to achieve the flipping, and subsequently, the table can be placed within the reachable range of the robot, as illustrated in Fig.~\ref{application}.
	Such scenarios involving flipping manipulations emphasize the significance of our planner in extending the capabilities of multiple mobile manipulator systems in complex tasks.

	\subsection{Result Discussion}

	These simulations and experiments correspondingly illustrate the contributions.
	To the best of our knowledge, similar manipulation to Fig.~\ref{application} has not been reported in other research yet.
	In related planners of MMMS \cite{zhang2021task,jiao2015transportation,alonso2017multi}, they mainly focus on object transportation and do not consider the problem of configuration connectivity.
	Compared to them, the proposed planner adapts to a broader range of scenarios.
	

	Both regrasping and platform motions improve manipulator connectivity but serve different functions.
	Platform motions expand the reachable space, allowing a single configuration to handle more object trajectories, resolving issues like joint limits or collisions.
	Regrasping, however, switches configurations without changing the reachable space.
	While platform motion may seem superior, regrasping is also essential.
	As shown in Fig.~\ref{simulation2}(b), the manipulation can not be completed without regrasping or platform motions.
	Since regrasping is often challenging in reality, especially with heavy objects, as robots may need to rely on environmental support to reposition objects.
	Therefore, the proposed planner uses regrasping only when necessary and minimizes its number.

	\subsection{Limitations of Proposed Framework}
	
	The results successfully demonstrate the effectiveness of our planning framework for multiple mobile manipulator systems in complex manipulations.
	Both regrasping and platform motions are employed to enhance the configuration connectivity under closed-chain constraints.
	Additionally, the proposed planner tries to accomplish the task with the minimum number of regrasping.
	Still, there are certain limitations within our planning framework warranting further investigation.

	\subsubsection{Completeness of Proposed Framework}
	
	In our planner, two specific optimization problems are introduced to ensure the configuration connectivity.
	However, the completeness of the framework has not been verified.
	When planning fails, it remains uncertain whether it is due to a genuinely unavailable path or an unreasonable planning method.
	
	\subsubsection{Trajectories of Object and Grasping Poses}
	
	The trajectory of the object and grasping poses are predetermined and considered as input to the planner.
	However, in some cases, it is only necessary to specify the initial pose and final pose of the object, and the specific trajectory or grasping poses are not required.
	As shown in Fig.~\ref{simulation2}, different trajectories of object may have a significant impact on the motions of robots.
	Similarly, different grasping poses can lead to different motions of robots and may even determine whether the manipulation can be successfully completed.
	
	\section{Conclusion and Future Work}
	
	This paper proposes a novel planning framework of multiple mobile manipulator systems in complex manipulation.
	The trajectory and grasping poses of object are considered as inputs of the planner and the outputs are motions of robots.
	The manipulation is mainly completed through motions of the manipulator.
	When manipulators suffer configuration disconnectivity, platform motions or regrasping are enrolled.
	Experimental results successfully demonstrate the effectiveness of our framework.
	
	Regarding future work, on the one hand, we will further refine the planning framework, for example, the completeness of the planner.
	Moreover, the framework can be extended to consider the trajectory of object and grasping poses as outputs, which may provide greater flexibility in planning.
	On the other hand, we will study the practical manipulation issues.
	For example, robust cooperative control strategies, which are crucial for ensuring safety in collaborative operations, will be further researched.

	\bibliographystyle{IEEEtran}
	\bibliography{References}

\end{document}